\documentclass[runningheads]{llncs}

\usepackage{eccv}

\usepackage{eccvabbrv}

\usepackage{graphicx}
\usepackage{booktabs}
\usepackage{esvect}
\usepackage{multirow}
\usepackage{bbding}
\usepackage{algorithm}
\usepackage{algorithmic}
\usepackage{wrapfig}

\usepackage[accsupp]{axessibility}  

\usepackage{hyperref}

\usepackage{orcidlink}

\begin{document}

\title{RISurConv: Rotation Invariant Surface Attention-Augmented Convolutions for 3D Point Cloud Classification and Segmentation} 

\titlerunning{RISurConv}

\author{Zhiyuan Zhang\inst{1}\orcidlink{0000-0003-3945-5638} \and
Licheng Yang\inst{2}\thanks{co-first author.} \and
Zhiyu Xiang\inst{2}\orcidlink{0000-0002-3329-7037}}

\authorrunning{Z.~Zhang et al.}

\institute{School of Computing and Information Systems, Singapore Management University 
\\
\url{https://github.com/cszyzhang/RISurConv}
\and
College of Information Science and Electronic Engineering, Zhejiang University}
\def\ourconv{RISurConv~}
\maketitle

\begin{abstract}
Despite the progress on 3D point cloud deep learning, most prior works focus on learning features that are invariant to translation and point permutation, and very limited efforts have been devoted for rotation invariant property. Several recent studies achieve rotation invariance at the cost of lower accuracies. In this work, we close this gap by proposing a novel yet effective rotation invariant architecture for 3D point cloud classification and segmentation. Instead of traditional pointwise operations, we construct local triangle surfaces to capture more detailed surface structure, based on which we can extract highly expressive rotation invariant surface properties which are then integrated into an attention-augmented convolution operator named \ourconv to generate refined attention features via self-attention layers. Based on \ourconv we build an effective neural network for 3D point cloud analysis that is invariant to arbitrary rotations while maintaining high accuracy. We verify the performance on various benchmarks with supreme results obtained surpassing the previous state-of-the-art by a large margin. We achieve an overall accuracy of \textbf{96.0\%} (+4.7\%) on ModelNet40, \textbf{93.1\%} (+12.8\%) on ScanObjectNN, and class accuracies of \textbf{91.5\%} (+3.6\%), \textbf{82.7\%} (+5.1\%), and \textbf{78.5\%} (+9.2\%) on the three categories of the FG3D dataset for the fine-grained classification task. Additionally, we achieve \textbf{81.5\%} (+1.0\%) mIoU on ShapeNet for the segmentation task.
\end{abstract}

\section{Introduction}
\label{sec:intro}

Point cloud has become the most promising 3D data representation for a wide range of immersive applications from robot navigation to autonomous driving. The increasing availability of 3D sensors and the emergence of large and mid-scale point cloud datasets ~\cite{armeni20163d,chang2015shapenet,wu-3dshapenets-cvpr15,hua2016scenenn,uy-scanobjectnn-iccv19,liu2021fine} have spurred significant research in this area, leading to the development of numerous deep learning models for point cloud classification and segmentation, which are the fundamentals for various computer vision tasks. 

However, analyzing 3D point clouds remains challenging, mainly due to the irregular nature of point clouds and their inherent invariances such as translation, point permutation, and rotation. While significant progress has been made in learning translation and permutation-invariant features~\cite{qi2017pointnet,qi2017pointnet++,li2018pointcnn,zhao2020quaternion,qian2022pointnext}, achieving rotation invariance in point cloud convolution has been a relatively unexplored area.

Rotation invariance is essential for 3D object classification and segmentation as objects can be viewed from different viewpoints, leading to variations in their orientation. Prior approaches~\cite{qi2017pointnet,qi2017pointnet++,wang2018edgeconv} usually rely on rotation augmentation in training stage to relief the rotation affections for testing. However, such scheme is less effective as 3D data has more degrees of freedom. A rotation invariant model is, therefore, critical to accurately classify and segment objects in 3D point clouds. 

Various attempts have been made to address the problem of rotation invariance in 3D point clouds, as evidenced by several works~\cite{zhang-riconv-3dv19,rao-spherical-cvpr19,poulenard-spherical-3dv19,deng2018ppf,chen2019clusternet}. These efforts have focused on designing rotation invariant properties to achieve consistent accuracies under arbitrary rotations without resorting to rotation augmentation. However, these methods have demonstrated inferior performance when compared to translation invariant approaches, primarily due to the loss of global information during the generation of rotation invariant properties. Recent research efforts~\cite{zhang2020global,kim2020rotation,thomas2020rotation} have attempted to overcome this limitation by employing local reference frame (LRF) or local reference axis (LRA)~\cite{zhang2022riconv++} to transform the data into a canonical coordinate system and encode global information. Despite the improvements, their accuracies are still lower than those of state-of-the-art non-rotation invariant methods because of the unstable LRF/LRA and the less descriptive rotation invariant properties that the useful surface information is not well preserved.

To address the above-mentioned issues, we propose a novel yet effective rotation invariant architecture. Specifically, we construct local triangle surfaces for each reference point of the input data to better capture the local surface structure. On the local surfaces, we design highly expressive rotation invariant surface properties which are then integrated into an attention-augmented convolution operator named \ourconv to extract refined rotation invariant features. Finally, we build up network based on \ourconv for rotation invariant object classification and segmentation. In summary, our main contributions include:

\begin{itemize}
\item \textbf{Rotation Invariant Surface Properties}. We construct local triangle surfaces from the reference point and its neighbors, based on which we are able to design highly expressive rotation invariant surface properties;
\item \textbf{RISurConv}. We integrate the Rotation Invariant Surface Properties into an attention-augmented architecture named \ourconv comprising two self-attention layers to learn and generate refined rotation invariant features; 

\item Extensive experiments on a variety of classification and segmentation tasks. Our approach shows supreme performance, \textbf{surpassing the state-of-the-art by a large margin} under challenging rotation scenarios including an analysis of rotation-invariant features and an ablation study of our neural network to provide insights into the key factors contributing to the exceptional performance.
\end{itemize}

\section{Related Works}
\label{related}
In this section, we review the representative works in 3D point cloud deep learning and rotation invariant learning.

\textbf{3D Point Cloud Deep Learning.}
3D point clouds is a more compact and intuitive representation for feature learning. 
PointNet~\cite{qi2017pointnet} pioneered a point cloud convolution with global features by max-pooling per-point features from MLPs, and follow-up works~\cite{qi2017pointnet++,li2018pointcnn,xu2018spidercnn} have focused on exploring convolution kernels that exploit geometric features~\cite{shen2018mining}, adding edges on top of points~\cite{wang2018edgeconv}, parameterizing convolution using polynomials~\cite{xu2018spidercnn}, and leveraging shape context~\cite{xie2018shapecontext}. Some methods are designed to combine with recurrent neural networks~\cite{huang2018recurrent} and sequence models~\cite{liu2018point2seq}. 

In recent years, attention mechanism has become increasingly popular in various natural language processing (NLP)~\cite{vaswani2017attention} and computer vision~\cite{dosovitskiy2020image} tasks. With their ability to handle sequential and spatial information, attention mechanism has also shown promise in 3D point cloud processing tasks. Pioneer works that apply attention mechanism in point clouds include Point Transformer~\cite{zhao2021point} and PCT~\cite{guo2021pct}, which use a self-attention mechanism to learn local and global features of point clouds. Other works have explored different types of transformers. Inspired by BERT~\cite{devlin2018bert}, Point-BERT~\cite{yu2022point} was proposed to pre-train pure Transformer-based models with a Mask Point Modeling task for point cloud classification. Dual Transformer Network~\cite{han2022dual} aggregated the point-wise and
channel-wise multi-head self-attention models simultaneously such that long-range context dependencies can be captured by investigating the point-wise and channel-wise relationships. Overall, these developments show promise in enhancing the efficiency and accuracy of point cloud analysis for a variety of applications. Readers can refer to the survey~\cite{guo-point-survey-2019} for a comprehensive overview of 3D point cloud deep learning. However, most of these methods lacks the property of rotation invariance which is critical for object classification and segmentation as the object can be viewed at any angle in reality. To address this issue, a common approach is to increase the training data by augmenting it with arbitrary rotations~\cite{qi2017pointnet++,wang2018edgeconv,li2018pointcnn}. 
However, such approach has difficulty in generalizing predictions to unseen rotations, resulting in deteriorated performance. Therefore, it would be desirable to design a specific convolution possessing rotation-invariant features.

\textbf{Rotation Invariant Learning.} 
Various methods have been proposed to address the issue of rotation invariance in feature learning of point clouds. For instance, Rao et al.~\cite{rao-spherical-cvpr19} used a spherical domain to define a rotation-invariant convolution for point clouds. However, the discretized sphere is sensitive to global rotations, which can lead to a notable drop in performance for objects with arbitrary rotations. To overcome this limitation, Poulenard et al.~\cite{poulenard-spherical-3dv19} integrated spherical harmonics to their convolution. Similarly, Chen et al.~\cite{chen2019clusternet} proposed a hierarchical clustering scheme to encode the relative angles between two-point vectors and used vector norm to maintain rotation invariance. Zhang et al.~\cite{zhang-riconv-3dv19} presented a simple convolution named RIConv that operates on handcrafted features built from Euclidean distances and angles that are rotation invariant by nature. However, this approach only considers local features resulting in accuracy degradation. GCAConv~\cite{zhang2020global} addressed this limitation by building a global context-aware convolution based on anchors and Local Reference Frame (LRF) to achieve rotation invariance. RI-GCN~\cite{kim2020rotation} learned rotation-invariant local descriptors and applied graph convolutional neural networks to aggregate local features based on LRF. Thomas~\cite{thomas2020rotation} also relied on LRF and used a multiple alignment scheme to achieve better results. RIF~\cite{li2021rotation} presented a framework to construct both local and global features based on distances, angles, and reference points. To remove the uncertainty of LRF in x and y directions,  RIConv++~\cite{zhang2022riconv++} extracts informative features based on local reference axis (LRA). However, the LRF/LRA and global reference used in these methods may not be stable enough limiting their overall performance. 
We can clearly see the accuracy discrepancy of the object classification task on ModelNet40 dataset~\cite{wu20153d}.
State-of-the-art translation-invariant convolutions such as PointNet++~\cite{qi2017pointnet++} or Point Transformer v2~\cite{wu2022point} achieve 89\%-94\% of accuracy while rotation-invariant convolutions only report
up to 87\%-91\% of accuracy~\cite{kim2020cycnn,zhang2022riconv++}.

Our target in this paper is not only closing this performance gap but also surpassing the state-of-the-art methods while maintaining rotation invariance.

\section{Rotation Invariant Surface Property}
\label{sec:risp}
In this section, we detail the Rotation Invariant Surface Property construction which is the first step of our method. Our goal for this step is to design highly expressive rotation invariant properties from underlying local surfaces. Different from previous works that rely on pointwise operations, we construct local surfaces around the reference point to better capture the local surface structure, based on which we then extract more expressive rotation invariant properties.

\begin{figure}[t]
\centering
\includegraphics[width=0.85\linewidth]{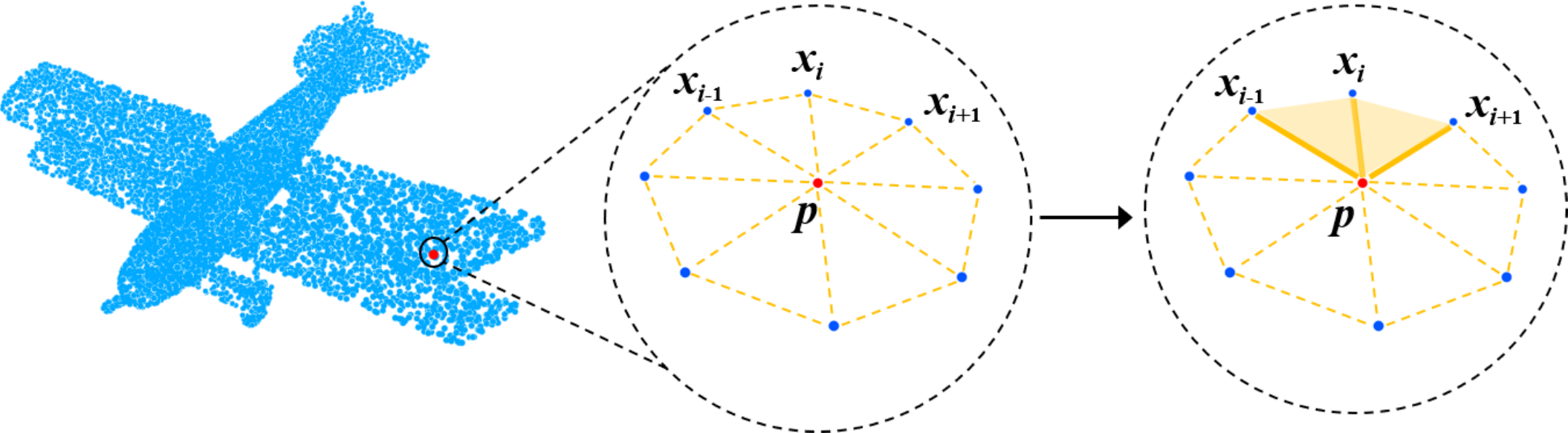}
\caption{Rotation Invariant Surface Property (RISP) construction: Given a point $p$ as the reference point, $K$ ($K = 8$ in this example) nearest points are selected (middle). 
For each neighbor $x_{i}$, two adjacent neighbors $x_{i-1}$ and $x_{i+1}$ are used to form two triangular local surfaces (right), based on which rotation invariant properties are constructed.
}
\label{fig:riproperty}
\end{figure}

The construction of rotation invariant surface property is shown in~\cref{fig:riproperty}. Given a point $p$ as the reference point (red), we get K nearest neighbors to form a local point set. In this case, $K$ is 8. 
For each neighbor $x_{i}$, two adjacent neighbors, $x_{i-1}$ and $x_{i+1}$, are identified based on the Euclidean distances, forming two triangular local surfaces, as shown in the figure. 
Then, we construct the rotation invariant surface properties (RISP) as follows:
\begin{equation}
\begin{split}
\label{eq:1}
	\mathrm{RISP}(x_i) = [L_{0}, \phi_{1}, \phi_{2}, \phi_{3}, \phi_{4}, \phi_{5}, \\ \alpha_{1}, \alpha_{2},  \beta_{1}, \beta_{2}, \theta_{1}, \theta_{2}, \gamma_{1}, \gamma_{2}]\,
\end{split}
\end{equation}
where $L_{0}$ measures the distance from reference $p$ to neighbor $x_{i}$, and  $\phi_{1}$ to $\phi_{5}$ measure the two triangles as well as the relationship between the two triangle surfaces with regard to the edge $\vv{px_{i}}$ in the Euclidean space:

\begin{align} 
\label{eq:2}
\phi_{1} &= \angle\left( \vv{x_{i-1}p} , \vv{x_{i}p} \right), \phi_{2} = \angle\left( \vv{x_{i+1}p}, \vv{x_{i}p} \right), \\
\phi_{3} &= \angle\left( \vv{x_{i-1}x_{i}}, \vv{x_{i-1}p} \right), \phi_{4} = \angle\left(  \vv{x_{i+1}p}, \vv{x_{i+1}x_{i}} \right), \nonumber\\
\phi_{5} &= \angle\left(  \vv{x_{i+1}p} \times \vv{x_{i}p}, \vv{x_{i-1}p} \times \vv{x_{i}p} \right), \nonumber 
\end{align}
while other properties describe the two surfaces in the tangent space, e.g., normal vectors can define the directions in which the surface is bending away from the tangent space:
\begin{align} 
\label{eq:3}
\alpha_{1} &= \angle\left( \vv{n_{p}} , \vv{x_{i}p} \right),  \alpha_{2} = \angle\left( \vv{n_{p}}, \vv{x_{i-1}p} \right),  \\ 
\beta_{1} &= \angle\left( \vv{n_{i}} , \vv{x_{i}p} \right), \beta_{2} = \angle\left( \vv{n_{i}}, \vv{x_{i-1}x_{i}} \right), \nonumber \\
\theta_{1} &= \angle\left( \vv{n_{i-1}} , \vv{x_{i-1}p} \right), \theta_{2} = \angle\left( \vv{n_{i-1}}, \vv{x_{i-1}x_{i}} \right), \nonumber \\
\gamma_{1} &= \angle\left( \vv{n_{i+1}}, \vv{x_{i+1}x_{i}} \right), \gamma_{2} = \angle\left( \vv{n_{i+1}}, \vv{x_{i+1}p}  \right). \nonumber
\end{align}

RISP is able to fully describe the dual triangles as well as their relationship along different directions. Readers can refer to the supplementary for the mathematical \textbf{proof of the completeness } of RISP defined in Equation~\ref{eq:1}.

\paragraph{Number of local triangle surfaces.} Throughout the paper, we define the number of local triangle surfaces as 2. Such setting is due to the fact that we wish to capture local surface as much as possible to extract more useful information to reflect the real shape of the object. So we construct 2 triangles rather than single triangle. However, this does not mean the more the better since more triangles can result in distorted surface and affect the computing efficiency. Please refer to the ablation study section for more details. 

\section{The \ourconv Operator}
\begin{figure}[htb]
\centering
\includegraphics[width=0.95\linewidth]{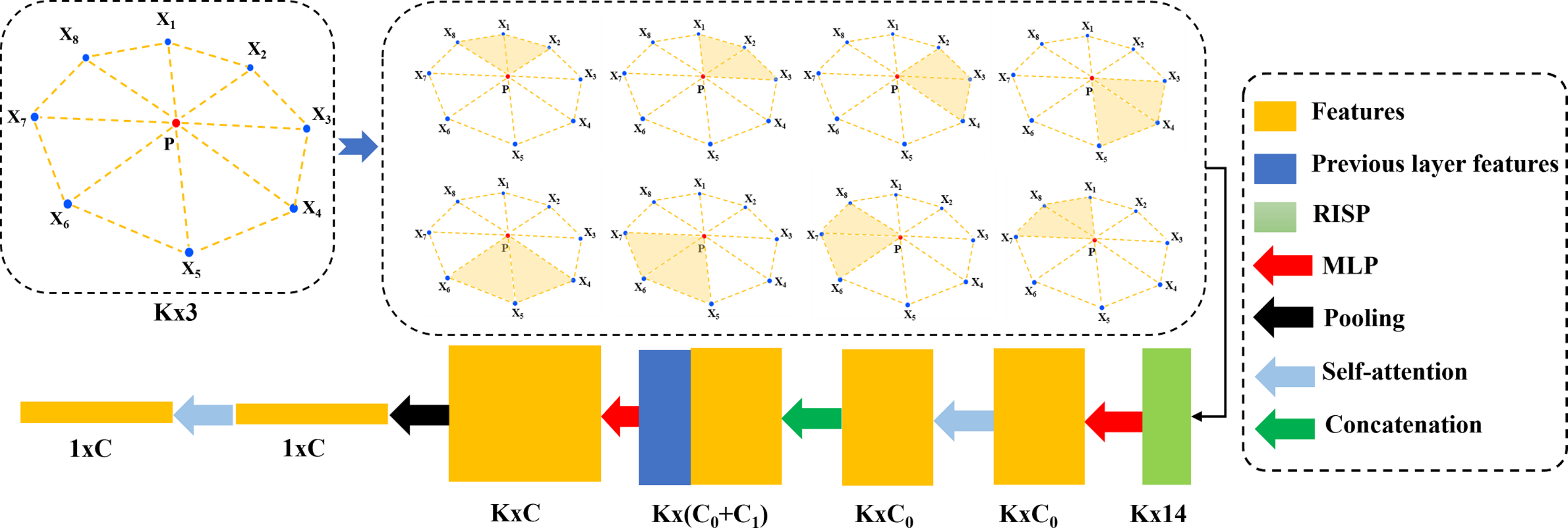}
\caption{\ourconv operator. For a local point set with $p$ as the reference (red), K nearest neighbors are labelled as blue. Then, we compute the Rotation Invariant Surface Properties at each neighbor by constructing local dual triangle surfaces (Section~\ref{sec:risp}), which is embedded to a high-dimensional space by a shared multi-layer perceptron (MLP) followed by a self-attention layer to produce refined features. 
Concatenated with previous layer features (if any), the features of these local points are further passed to MLPs, which are then summarized by maxpooling. To further refine the features, another self-attention layer follows.}
\label{fig:ourconv}
\end{figure}

\begin{algorithm}[tb]
	\small
	\caption{\ourconv operator.}
	\label{alg:conv}
	\textbf{Input}: Reference point $p$, point set $\Omega$, point features $\mathbf{f}_{prev}$ from 	previous layer (if any)\\
	\textbf{Output}: Convoluted features $\mathbf{f}$
	\begin{algorithmic}[1]
		\STATE $\mathbf{f} \leftarrow \left\{ \mathrm{RISP}(x_i) : \forall x_i \in \Omega \right\}$ \hfill * Construct Rotation Invariant Surface Properties (Section~\ref{sec:risp})
		\STATE $\mathbf{f} \leftarrow \mathrm{MLP}(\mathbf{f})$; \hfill * Embed each feature to a high-dimensional feature space
		\STATE $\mathbf{f} \leftarrow \mathrm{SA}(\mathbf{f})$; \hfill * Refine features via self-attention layer 
		\STATE $\mathbf{f}_{in} \leftarrow [\mathbf{f}_{prev}, \mathbf{f}]$ \hfill * Concatenate the features from the local and the previous layer (if any)
		\STATE $\mathbf{f}_{out} \leftarrow \mathrm{MLP}(\mathbf{f}_{in})$ \hfill *  Feature embedding
		\STATE $\mathbf{f}_{out} \leftarrow \mathrm{maxpool}(\mathbf{f}_{out})$ \hfill *  Maxpool features\\
		\textbf{return} $\mathrm{SA}(\mathbf{f}_{out})$ \hfill * Self-attention and return
	\end{algorithmic}
\end{algorithm}

Based on the Rotation Invariant Surface Properties (RISP), we are able to build up RISurConv. The main steps are detailed in~\cref{fig:ourconv}. To start, we utilize a farthest point sampling strategy to generate uniformly distributed representative points. From this initial set, we perform K-nearest neighbor searches to obtain local point sets. Let us denote a local point set by $\Omega = \{x_i\}$, where $x_i$ represents the 3D coordinates of point $i$. We define the attention-augmented convolution operation as follows: 
\begin{align} 
\mathbf{f}(\Omega) = SA( \sigma( \mathcal{A} ( \{ \mathcal{T}(\mathbf{f}_{x_i}) : \forall i \} ) ) ).
\end{align}
This formula indicates that features of each point in the point set are first transformed by $\mathcal{T}$ before being aggregated by the aggregation function $\mathcal{A}$ and passed to an activation function $\sigma$. SA is a channel-wise self-attention layer used to output refined features. 
We set the input features to our expressive rotation-invariant features $\mathbf{f}_{x_i} = \mathrm{RISP}(x_i)$.
We define the transformation function as
\begin{align}
\mathcal{T}(\mathbf{f}_{x_i}) = \mathbf{w}_i \cdot \mathbf{f}_{x_i} = \mathbf{f}'_{x_i}
\end{align}
where $\cdot$ indicates the element-wise product, and $\mathbf{w}_i$ is the weight parameter to be learned by the network. 
Our transformation function is similar to PointNet++.
A popular choice of the aggregation function $\mathcal{A}$ is maxpooling, which supports permutation invariance  of the input point features.

To proceed with feature learning and be invariant to point permutation, we construct surfaces for each neighbor from which we construct RISPs. In~\cref{fig:ourconv}, there are eight neighbor, so the resulted RISPs are in size of $8\times 14$. Here, $14$ is the length of the RISP as defined in~\cref{eq:1} and $8$ is the neighborhood size. Since RISPs are used as the input which are already rotation invariant, \ourconv is rotation invariant by nature. We then embed the RISPs into feature space using two layers of MLPs followed by a self-attention (SA) layer for feature refinement. We concatenate the refined features with previous layer features (if any), and embed the concatenated features into higher dimensional space via MLPs again. After the maxpooling, the aggregated features are passed into another SA layer again for further refinement. 
The detailed steps are shown in the~\cref{alg:conv}. Here, the two SA layers used in \ourconv are the standard
SA module of the transformer~\cite{vaswani2017attention}. 

\textbf{Self Attention Layers}.
\label{sa}
In \ourconv there are two Self-Attention (SA) layers by which the extracted features are enhanced. Below, we provide a more detailed illustration of the structure of the SA layers. The SA structure is the standard SA module of the famous transformer~\cite{vaswani2017attention} (Attention is all you need) which runs twice in RISurConv. The first SA is for the K neighbors, and the second SA works for the N representative points. The input tensor shape is $[B,N,K,C_{0}]$ for the first SA which goes into linear layers and outputs $Q$,$K$, and $V$ respectively in shape of $[B,N,K,C_{0}]$. According to $attn(Q,K,V)=Softmax(QK^\top/sqrt(d_{k}))V$, we have the 1st attention score in $[B,N,K,K]$. By multiplying with V, we get the refined feature in $[B,N,K,C_{0}]$. The same goes with the second SA where the input shape is $[B,N,C]$ and the attention score shape is $[B,N,N]$.

\section{\ourconv Networks}
\begin{figure}[ht]
\centering
\includegraphics[width=0.98\linewidth]{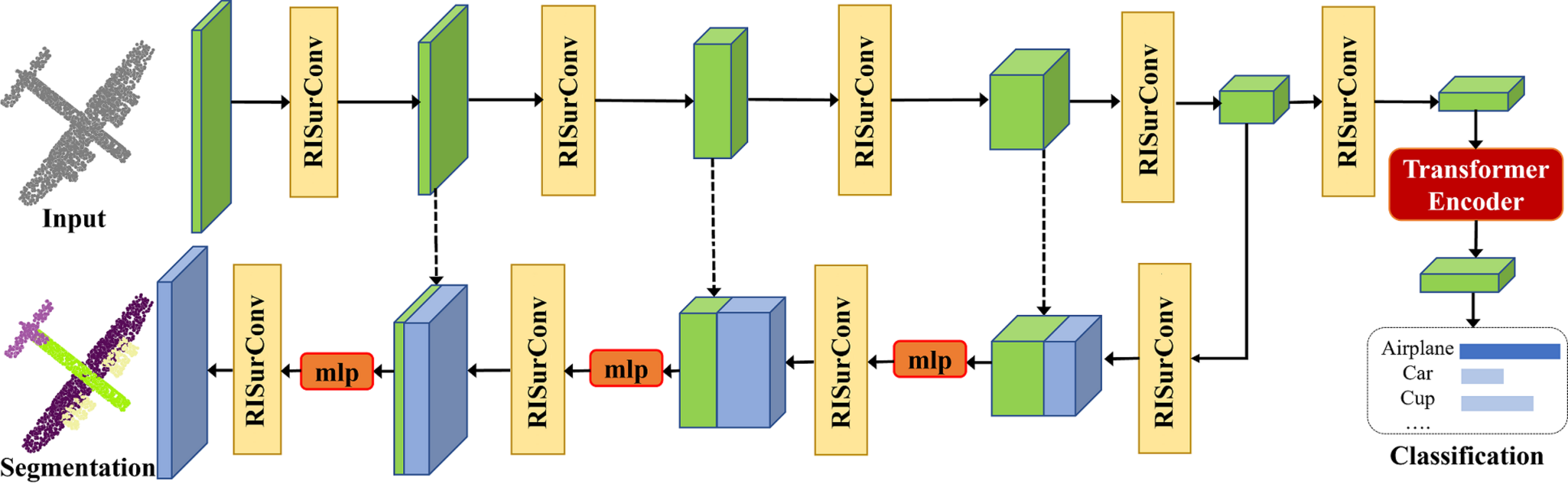}
\caption{Our neural network architecture comprises five \ourconv layers to extract rotation invariant features followed by a Transformer Encoder to enhance the learnt features before fully connected layers for object classification. We add a decoder with skip connections for segmentation task.}
\label{fig:network}
\end{figure}
We employed the \ourconv to develop neural networks for object classification and segmentation, as illustrated in~\cref{fig:network}. Our classification network follows a standard architecture similar to the single-scale grouping version of PointNet++, consisting of five consecutive layers of \ourconv followed by a transformer encoder~\cite{vaswani2017attention} with 8 heads to enhance the extracted features. Finally, we connect the network to fully connected layers to produce the probability map. One of the major advantages of \ourconv is its ability to handle arbitrary rotation and point orders, which enables us to place each \ourconv layer consecutively without the need for complex preprocessing steps. Additionally, we apply batch normalization and ReLU activation to each convolution layer by default. The segmentation network utilizes an encoder-decoder structure similar to the U-Net~\cite{ron2015unet}. We adopt the same definition of deconvolution as RISurConv. The key distinction lies in that the convolution produces a point subset with a greater number of feature channels and less points while the deconvolution outputs to a point set with more points compared to the input with fewer feature channels. Please refer to the supplementary for more details regarding the network architecture.

\section{Experiments}
We report our evaluation results in this section. Our network is implemented in PyTorch using an Adam optimizer with initial learning rate of 0.001 for both classification and segmentation. Our training is executed on a computer with an Intel(R) Core(TM)
i7-10700K CPU equipped with a NVIDIA GTX 3090 GPU.

We evaluated the performance of our method extensively on a series of classification and segmentation benchmarks. To train our model, we utilized 450 epochs for classification and 250 epochs for segmentation for most datasets except the Car and Chair categories of the fine-grained FG3D dataset due to small size and subtle variation where we applied 900 epochs. Our method can achieve convergence within 11 hours for classification and 24 hours for segmentation. To assess the robustness of our method, we followed the experimental design proposed by~\cite{esteves2019equivariant} and conducted experiments in three cases.
The first case involved training and testing with data augmented with rotation about the gravity axis (z/z), which is a common approach for evaluating translation-invariant point cloud learning methods. The second case involved training and testing with data augmented with arbitrary SO3 rotations (SO3/SO3), and the third case involved training with data by z-rotations and testing with data by SO3 rotations (z/SO3), which are used to evaluate rotation invariance. A rotation-invariant method should produce consistent results across all three cases.
Our method demonstrated superior performance across all three cases, highlighting its robustness and effectiveness.

\subsection{Human-made Object Classification}
ModelNet40~\cite{wu20153d} that comprises 9843 training models and 2468 test models, divided into 40 categories is one the of the most popular datasets for point cloud classification evaluation. The input data size is 1024 with each point equipped with (x; y; z; nx; ny; nz) which are 3D coordinates and 3D normals in the Euclidean space. Note that \textbf{normals are optional} to our method. When normals are not available (w/o n), we use weighted eigen vector corresponding to the smallest eigen value as the normal. This strategy has been used in SHOT and RIConv++~\cite{zhang2022riconv++} which is highly robust to noise and efficient to compute.

The experimental results are shown in~\cref{tab:classification_modelnet40}. We use two criteria for evaluation: overall accuracy and
accuracy standard deviation (Std.). 
True rotation invariant methods are expected to be unaffected by the rotations present in the training and testing data, resulting in low accuracy deviation.

From~\cref{tab:classification_modelnet40}, we see that our method surpasses the state-of-the-art performance in all cases while maintaining zero accuracy deviation. It is worth noting that our method surpasses both rotation invariant and non-rotation invariant  methods. To the best of our knowledge, this is the first work that achieves such high accuracy. It outperforms the state-of-the-art rotation invariant method RIConv++ by \textbf{4.7\%}, and outperforms the state-of-the-art non-rotation invariant method Point Transformer v2 by \textbf{1.8\%} under z/z case.
 
\begin{table}[htb]
\caption{Comparisons of the classification accuracy (\%) on the ModelNet40 dataset. On average, our method has the best accuracy and lowest accuracy deviation in all cases (\textbf{Overall Accuracy}).}
\centering
\label{tab:classification_modelnet40}
 \scalebox{0.9}{
\begin{tabular}{l|l|llr|ccc|cc}
\toprule
& Method  & Format & Input Size & Params.& z/z$\uparrow$ & SO3/SO3$\uparrow$ &z/SO3$\uparrow$ & Std.$\downarrow$  \\
\midrule
\multirow{7}{*}{\rotatebox[origin=c]{90}{Traditional}} & VoxNet~\cite{maturana2015voxnet} & voxel & $30^{3}$ &0.90M & 83.0 & 87.3 & - & 3.0 \\
& SubVolSup~\cite{qi2016volumetric}& voxel & $30^{3}$ &17.00M  & 88.5 & 82.7 & 36.6 & 28.4 \\ 
		& PointNet~\cite{qi2017pointnet}  & xyz & $1024\times 3$ & 3.50M & 87.0 & 80.3 & 21.6 & 41.0 \\ 
		& PointCNN~\cite{li2018pointcnn}  & xyz & $1024\times 3$ & 0.60M & 91.3 & 84.5 & 41.2 & 27.2 \\
		& PointNet++~\cite{qi2017pointnet++}  & xyz + nor & $1024\times 6$ & 1.40M & 89.3 & 85.0 & 28.6 & 33.8\\
		& DGCNN~\cite{wang2018edgeconv}  & xyz & $1024\times 3$ & 1.84M & 92.2 & 81.1 & 20.6 & 38.5 \\
		& RS-CNN~\cite{liu2019relation}  & xyz & $1024\times 3$ & 1.41M & 90.3 & 82.6 & 48.7 & 22.1\\
		& Pt Transformer~\cite{zhao2021point}  & xyz & $1024\times 3$ & - & 93.7 & 85.9 & 50.1 & 19.1\\		
		& Pt Transformer v2~\cite{wu2022point}  & xyz & $1024\times 3$ & - & 94.2 & 88.3 & 51.8 & 23.0\\	
\midrule
\multirow{12}{*}{\rotatebox[origin=c]{90}{Rotation-invariant}} & Spherical CNN~\cite{esteves2018learning}  & voxel & $2\times 64^{2}$ & 0.50M & 88.9 & 86.9 & 78.6 & 5.5\\
& RIConv ~\cite{zhang-riconv-3dv19} & xyz & 1024 $\times 3$ & 0.70M & 86.5 & 86.4 & 86.4 & 0.1\\
& SPHNet ~\cite{poulenard-spherical-3dv19} & xyz & 1024 $\times 3$ &  2.90M & 87.0 & 87.6 & 86.6 & 0.5 \\
& SFCNN~\cite{rao-spherical-cvpr19} & xyz & 1024 $\times 3$ &  - & 91.4 & 90.1 & 84.8 & 3.5 \\
& ClusterNet~\cite{chen2019clusternet} & xyz & 1024 $\times 3$ &  1.40M & 87.1 & 87.1 & 87.1  & \textbf{0.0} \\
& GCAConv~\cite{zhang2020global} & xyz & 1024 $\times 3$ & 0.41M & 89.0 & 89.2 & 89.1 & \textbf{0.0}\\
& RIF~\cite{li2021rotation} & xyz & 1024 $\times 3$ & - & 89.4 & 89.3 & 89.4 & \textbf{0.0}\\

& RI-GCN~\cite{kim2020rotation} & xyz + nor & 1024 $\times 6$ & 4.38M & 91.0 & 91.0 & 91.0 & \textbf{0.0}\\
& RIConv++~\cite{zhang2022riconv++}  & xyz & 1024 $\times 3$ & 0.40M & 91.2 & 91.2 & 91.2 & \textbf{0.0}\\
& RIConv++~\cite{zhang2022riconv++}  & xyz + nor & 1024 $\times 6$ & 0.40M & 91.3 & 91.3 & 91.3 & \textbf{0.0}\\
\midrule
& Ours (w/o normal)  & xyz & 1024 $\times 3$ & 14.0M & \underline{95.6} & \underline{95.6} & \underline{95.6} & \textbf{0.0}\\
& Ours (w/ normal) & xyz + nor & 1024 $\times 6$ & 14.0M & \textbf{96.0} & \textbf{96.0} & \textbf{96.0} & \textbf{0.0}\\
\bottomrule
\end{tabular}}
\end{table}

\subsection{Real World Object Classification}
\setlength\intextsep{-10pt}
\begin{wraptable}[14]{r}{0.55\textwidth}
\caption{Comparisons of real world 3D point cloud classification on hardest variant of ScanObjectNN dataset (\textbf{Overall Accuracy}).}\label{tab:classification_scanobjectnn}
\centering
	\scalebox{0.9}{
	\begin{tabular}{l|ccc}
		\toprule
		       &        &PB\_T50\_RS   &\\
		Method      &z/z  &SO3/SO3 &z/SO3\\
		\midrule
		PointNet~\cite{qi2017pointnet}      &68.2 &42.2 &17.1\\
		PointNet++~\cite{qi2017pointnet++}   &77.9 &60.1 &15.8\\
		PointCNN~\cite{li2018pointcnn}      &78.5 &51.8 &14.9\\
		DGCNN~\cite{wang2018edgeconv}      &78.1 &63.4 &16.1\\
		\midrule
		RIConv~\cite{zhang-riconv-3dv19}   &68.1 &68.3 &68.3\\
		GCAConv~\cite{zhang2020global}      &69.8 &70.0 &69.8\\
		RIConv++~\cite{zhang2022riconv++}   &80.3 &80.3 &80.3\\
		\midrule
		Ours (w/o normal) &\textbf{93.1} &\textbf{93.1} &\textbf{93.1}\\
		\bottomrule
	\end{tabular} }
\end{wraptable}
We also evaluate the classification performance on a real-world 3D point cloud dataset ScanObjectNN~\cite{uy-scanobjectnn-iccv19}. It is composed of 2902 point clouds categorized into 15 categories sampled from real world indoor scenes. For our evaluation, we use the processed files and choose the hardest variant PB\_T50\_RS with 50\% bounding box translation, rotation around the gravity axis, and random scaling that result in rotated and partial data. The results are shown in~\cref{tab:classification_scanobjectnn}. 
We see that \ourconv surpasses all the compared methods by a large margin. Particularly, our method \textbf{significantly outperforms} the state-of-the-art rotation invariant approach RIConv++~\cite{zhang2022riconv++} by \textbf{12.8\%}. This verifies that our method is effective for both synthetic and real world data. Note that we only test the ’w/o normal’ case as the normal vectors are not provided in the processed files of this dataset.

\subsection{Fine-Grained Object Classification}
\begin{table}[tb]
	\caption{Comparisons of fine-grained 3D point cloud classification on FG3D dataset (\textbf{Class Accuracy}).}
    \label{tab:classification_fg3d}
	\centering
	\scalebox{0.9}{
	\begin{tabular}{l|c|ccc|ccc|ccc}
		\toprule
		&       &         & Airplane      &    &    &Chair   &     &        &Car  &\\
		Method  &Modality &z/z  &SO3/SO3 &z/SO3  &z/z  &SO3/SO3 &z/SO3      &z/z  &SO3/SO3 &z/SO3\\
		\midrule
		VoxNet     &Voxel   &79.2 &57.5 &28.7 &73.3 &54.7 &16.7 &68.2 &42.2 &17.1\\
		\midrule
		MVCNN      &View &82.6 &66.9 &21.9  &76.3 &63.7 &14.6 &71.9 &51.8 &14.9\\
		View-GCN    &View &87.4 &56.6 &22.7 &79.7 &50.2 &19.7 &73.7 &44.4 &18.1\\
		RotationNet &View &89.1 &50.8 &29.8 &78.5 &36.2 &18.4 &72.5 &45.3 &28.3\\
		Part4Feature    &View &82.6 &49.1 &32.2 &77.1 &52.2 &30.2 &73.4 &35.0 &23.8\\
		FG3D-Net       &View &89.4 &61.1 &26.9 &80.0 &43.2 &18.4 &74.0 &56.4 &20.8\\
		\midrule
		PointNet     &Point &82.7 &55.5 &29.7 &72.1 &33.9 &18.2 &68.1 &32.1 &16.4\\
		PointNet++ &Point &87.3 &57.1 &25.6 &78.1 &40.4 &16.0 &70.3 &50.4 &20.8\\
		RS-CNN      &Point &82.8 &45.9 &32.2  &75.1 &50.7 &18.6 &71.2 &31.8 &13.9\\
		DGCNN     &Point &88.4 &60.6 &22.7 &71.7 &52.3 &19.7 &65.3 &53.4 &18.1\\
		RIConv  &Point &85.8 &85.8 &85.8 &76.3 &76.3 &76.3 &67.3 &67.3 &67.3\\
		RIConv++  &Point &87.9 &87.9 &87.9 &77.6 &77.6 &77.6 &69.3 &69.3 &69.3\\
		\midrule
		Ours &Point &\textbf{91.5} &\textbf{91.5} &\textbf{91.5} &\textbf{82.7} &\textbf{82.7} &\textbf{82.7} &\textbf{78.5} &\textbf{78.5} &\textbf{78.5}\\
		\bottomrule
	\end{tabular} }
\end{table}
Fine-grained object classification is more challenging because the differences between subcategories are subtle and require a high level of precision in distinguishing them. We conduct experiments on the three categories from FD3D dataset~\cite{liu2021fine}: Airplane, Chair and Car. To the best of our knowledge, this is the first work to test the performance of rotation invariant methods on fine-grained 3D point cloud dataset.
The class accuracies are shown in~\cref{tab:classification_fg3d}. Again, our method outperforms the  state-of-the-art approaches by large margins for all rotation cases and all the categories. Specifically, our method improves non-rotation invariant method FG3D-Net by \textbf{2.1\%}, \textbf{2.7\%}, and \textbf{4.5\%} for Airplane, Chair, and Car categories respectively under the z/z rotation case, and the accuracy consistency is well preserved. Compared to the rotation invariant methods, our method outperforms the latest RIConv++ by \textbf{3.6\%}, \textbf{5.1\%}, and \textbf{9.2\%} on the three categories. Such supreme performance shows that features extracted by \ourconv are not only rotation invariant but also highly expressive. 

\subsection{Part Segmentation on ShapeNet}
\begin{wraptable}[20]{r}{0.57\textwidth}
\caption{Comparisons of object part segmentation performed on ShapeNet dataset. The mean per-class IoU (mIoU, \%) is used to measure the accuracy under two challenging rotation modes: SO3/SO3 and z/SO3.}
\label{tab:partsegment}
\centering
 \scalebox{0.9}{
\begin{tabular}{l|c|cc} 
\toprule
Method   & Input & SO3/SO3 & z/SO3  \\
\midrule
PointNet~\cite{qi2017pointnet}     & xyz        & 74.4   & 37.8  \\
PointNet++~\cite{qi2017pointnet++} & xyz+nor & 76.7   & 48.2  \\
PointCNN~\cite{li2018pointcnn}     & xyz        & 71.4   & 34.7  \\
DGCNN~\cite{wang2018edgeconv}      & xyz        & 73.3   & 37.4  \\
RS-CNN~\cite{liu2019relation}      & xyz        & 72.5   & 36.5  \\
SpiderCNN   & xyz+nor & 72.3   & 42.9  \\
\midrule
RIConv~\cite{zhang-riconv-3dv19}  & xyz        & 75.5   & 75.3  \\
GCAConv~\cite{zhang2020global}     & xyz        & 77.3   & 77.2  \\
RI-GCN~\cite{kim2020rotation}      & xyz       & 77.0   & 77.0 \\
RIF~\cite{li2021rotation}    & xyz  & 79.4  & 79.2\\
RIConv++~\cite{zhang2022riconv++}   & xyz  & 80.3  & 80.3\\
RIConv++~\cite{zhang2022riconv++}   & xyz+nor  & 80.5  & 80.5\\
\midrule
Ours (w/o normal)          & xyz &  \underline{81.3} &  \underline{81.3} \\
Ours (w/ normal)           & xyz+nor &  \textbf{81.5} &  \textbf{81.5} \\
\bottomrule
\end{tabular}}
\end{wraptable}

To evaluate the segmentation performance, we employed the ShapeNet dataset comprising 16880 CAD models distributed across 16 distinct categories. These models are annotated with 2 to 6 parts each, culminating in a dataset of 50 object parts. We follow the standard train/test split with 14006 models for training and 2874 models for testing, respectively.

\begin{figure}[t]
\centering
\includegraphics[width=0.88\linewidth]{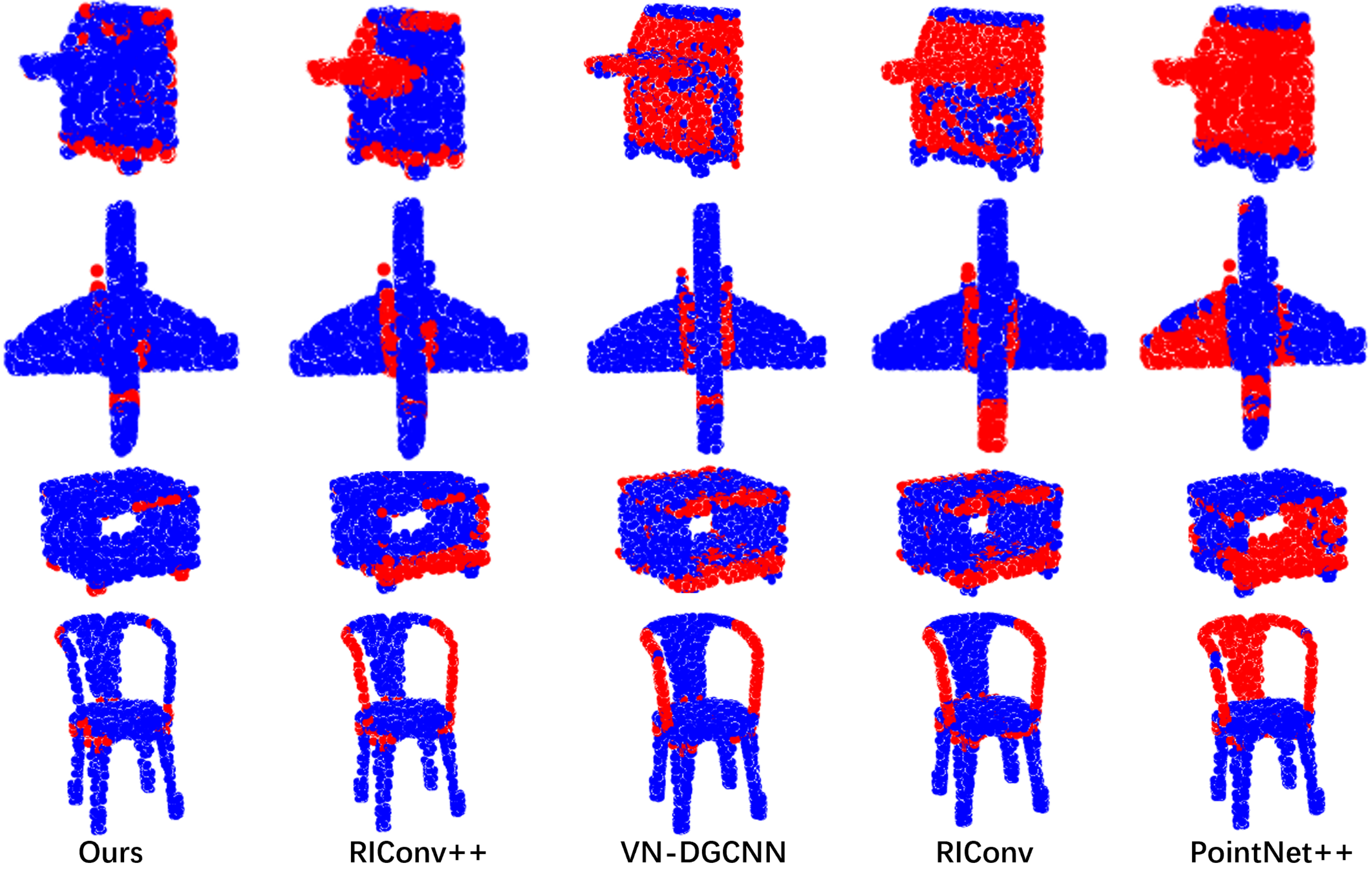}
\caption{Qualitative comparisons (Red indicates wrong). }
\label{fig:quality}
\end{figure}

The evaluation results are shown in~\cref{tab:partsegment}. Our method outperforms both traditional translation-invariant and latest rotation invariant methods significantly in the SO3/SO3 and z/SO3 scenario. Our method outperforms the RIConv++ by \textbf{1.0\%} mIoU. This result aligns well with the performance reported in the object classification task. 
We also show the qualitative results by error maps in~\cref{fig:quality}. The wrong segmentation points are plotted as red. It clearly shows that our predictions are the closest to the ground truth.

\subsection{Ablation Studies}
\label{ablation}
We ablate some vital designs of our method on ModelNet40~\cite{wu20153d} for an insightful exploration.
\begin{table}[b]
	\centering
	\begin{minipage}[t]{0.47\linewidth}
	\centering
\caption{Ablation study on the RISP design.}
	\label{tab-ablation}

	\scalebox{0.9}{
	\begin{tabular}{l|cc|c|c|c}
		\toprule
		Model  &$L_{0}$ & $\phi_{1...5}$  &$\alpha_{1,2}$  $\beta_{1,2}$  $\theta_{1,2}$  $\gamma_{1,2}$ &$\lambda$ $\mu$  &Acc.\\
		\midrule
		A     &\Checkmark    &\Checkmark   &\Checkmark  &  &96.0\\
		B     &    &\Checkmark  &\Checkmark  &  &95.5\\
		C     &    &   &\Checkmark  &  &90.9\\
		D     &\Checkmark    &\Checkmark &  &  &88.2\\
		E     &\Checkmark    &\Checkmark  &\Checkmark  & \Checkmark &95.7\\
		\bottomrule
	\end{tabular} }
	\end{minipage}
	\hfill
	\begin{minipage}[t]{0.47\linewidth}
	\caption{Ablation study on the Self-Attention module.}
	\label{tab:ablation2}
	\scalebox{0.9}{
		\begin{tabular}{l|ccc|c}
		\toprule
		Model  &SA1 & SA2  &Transformer Encoder  &Acc.\\
		\midrule
		A     &\Checkmark    &\Checkmark  &\Checkmark  &96.0\\
		B     &    &\Checkmark &\Checkmark  &95.6\\
		C     & \Checkmark    & &\Checkmark  &95.2\\
		D     &\Checkmark    &\Checkmark &   &94.3\\
		E     &    &   &   &92.8\\
		\bottomrule
	\end{tabular}}
	\end{minipage}
\end{table}

\textbf{RISP Design}. We first test the performance by turning on/off and adding different rotation invariant components used in the RISP construction (Algorithm~\ref{alg:conv}). 
The results are shown in~\cref{tab-ablation}. 
Model A is our baseline setting with all rotation invariant features activated.
Model B has only angle features but still achieves relatively high accuracy. Model C turns off both $L_{0}$ and $\phi_{1...5}$ with accuracy decreases to 90.9\%. This means that the distance $L_{0}$ is not as important as angles. 
In Model D, we turn off the angle features on the tangent direction with only $L_{0}$ and $\phi_{1...5}$ are kept which is more like RIConv~\cite{zhang-riconv-3dv19}.  
Compared to Model A, it shows that our proposed features is more effective than those by RIConv, and the improvement is explained by the additional consideration of the relations among the neighbor points. Model E adds more rotation invariant features. $\lambda$ and $\mu$ represent the angles $\angle\left( \vec{n_{p}} , \vec{n_{i+1}} \right)$ and $\angle\left( \vec{n_{p}} , \vec{n_{i-1}} \right)$ respectively. The results show that the accuracy does not increase. This verifies the completeness of the proposed RISP.

\textbf{Self-Attention Effects}.
We employ multiple self-attention (SA) modules to help produce refined rotation invariant features. So it is necessary to analyse the effects of different SA modules. In RISurConv, there are two SA modules and we name them as SA1 and SA2 respectively. Before fully connected layers, there is a Transformer Encoder module. We test these three modules by turning off one of them each time. From~\cref{tab:ablation2}, we see that when removing the first SA module in RISurConv, the performance drops a bit, while removing the second SA module also decreases the accuracy. This indicates that both SA1 and SA2 are important for the success of our method. In addition, we also test the performance when Transformer Encoder is removed. The overall accuracy decreases to 94.3\% but it is still higher than most existing methods. In model E, we test the performance when all SA modules are removed. The accuracy decreases to 92.8\% which is still better than the state-of-the-art rotation invariant method~\cite{zhang2022riconv++} thanks to the well designed RISPs which capture sufficient surface structures.  
\setlength\intextsep{10pt}
\begin{wrapfigure}[12]{r}{0.6\textwidth}
\centering
\includegraphics[width=0.6\textwidth]{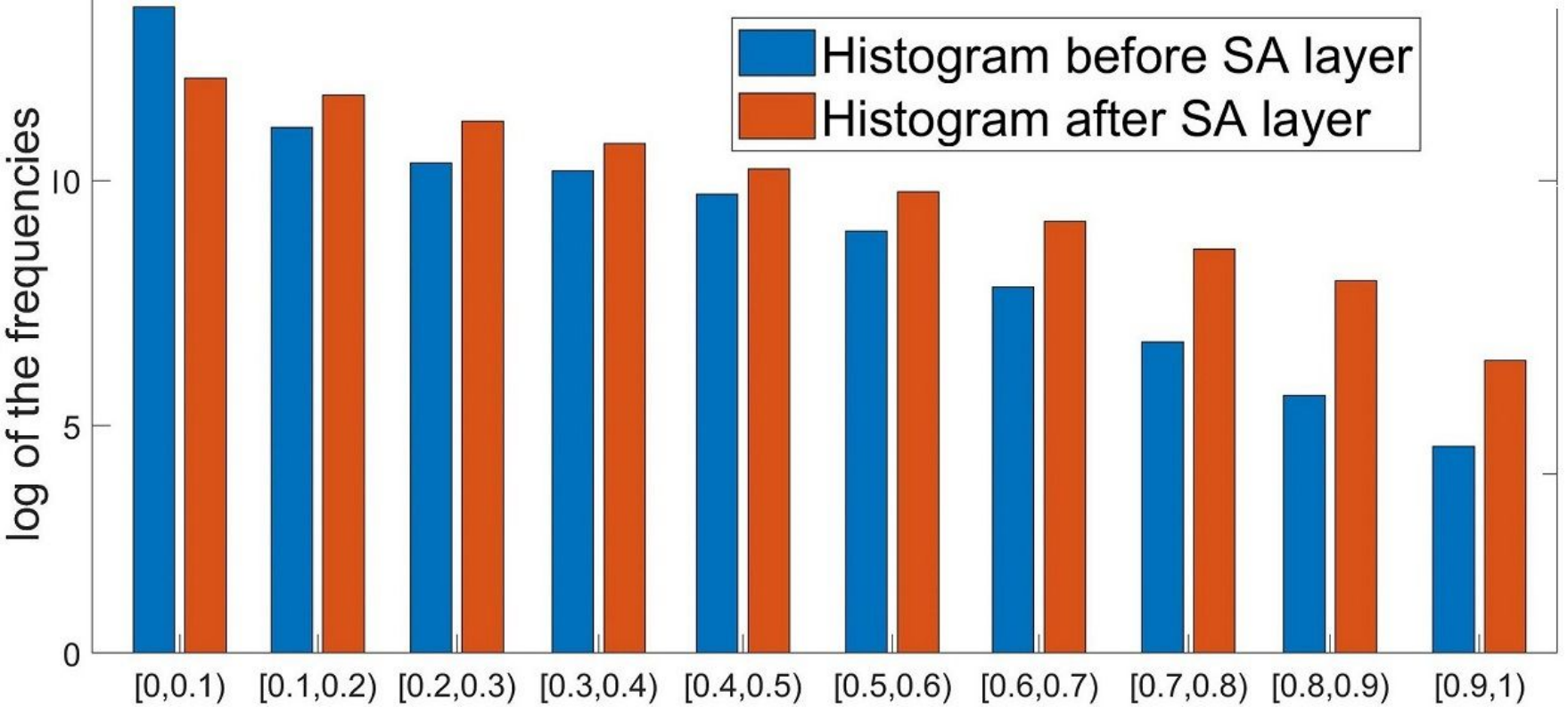}
\caption{Histogram comparison for normalized feature values without and with self-attention layers.}
\label{fig:sa}
\end{wrapfigure}
We also visualize the global feature vector before classification as shown in~\cref{fig:sa}. We normalize the feature values to $[0,1]$ and plot the histograms. We can see that by turning on the SA modules, the feature values become more evenly distributed and have less zeros because SA mechanism can reweight the features making the features more effective.

\setlength\intextsep{-10pt}
\begin{wraptable}[11]{r}{0.63\textwidth}
\caption{Trainable parameters, FLOPs and timing comparisons on ModelNet40. All are tested on the same platform. Time is measured per batch (batch size=16).}
\label{tab:parameters}
\centering
  \scalebox{0.9}{
    \begin{tabular}{l|c|c|c}
     \toprule
      Methods &Params & FLOPs & Time (Train/Infer)\\
      \midrule
      PointNet++~\cite{qi2017pointnet++} & 1.41M & 0.86G & 0.145s / 0.129s \\
    
      RIConv~\cite{zhang-riconv-3dv19} & 0.70M & 0.92G & 0.121s / 0.092s \\
      
      RI-GCN~\cite{kim2020rotation} & 4.19M & 1.24G & 0.148s / 0.112s \\

      RIConv++~\cite{zhang2022riconv++}  & 0.42M & 0.72G & 0.074s / 0.029s \\
      \midrule
      Ours & 13.96M & 1.12G & 0.114s / 0.049s \\
      Ours w/o TR & 1.35M & 1.11G & 0.105s / 0.047s \\
      \bottomrule
    \end{tabular} }
\end{wraptable} 
\textbf{Network Efficiency}.
In the experiments, we acknowledge that our method employs a higher number of parameters. Thus, it is imperative to conduct a comprehensive analysis of network efficiency during both the training and testing stages. We list the trainable parameters, FLOPs and running time comparisons in~\cref{tab:parameters}. We compare the performance with four recent works: PointNet++~\cite{qi2017pointnet++}, RIConv~\cite{zhang-riconv-3dv19}, RI-GCN~\cite{kim2020rotation}, RIConv++~\cite{zhang2022riconv++}.

From the results, we can see that the number of parameters mainly comes from the transformer architectures used in RISurConv. When we turn off all the transformer component, our method only takes up 1.35M parameters. 
More parameters do not mean longer inference time. It depends on lots of
things including depth, parameter, number of operations etc. Our approach
runs faster than most existing methods. It takes more time
for training because more epochs are needed for self attention layers to converge. This is a common phenomenon of transformer based methods. The accuracy is also not due to the complexity model. As is shown in the ablation studies, by removing transformer encoder, it still achieves state-of-the-art performance due to the effectiveness of RISurConv.
\setlength\intextsep{10pt}
\begin{wrapfigure}[11]{r}{0.38\textwidth}
\centering
\includegraphics[width=0.36\textwidth]{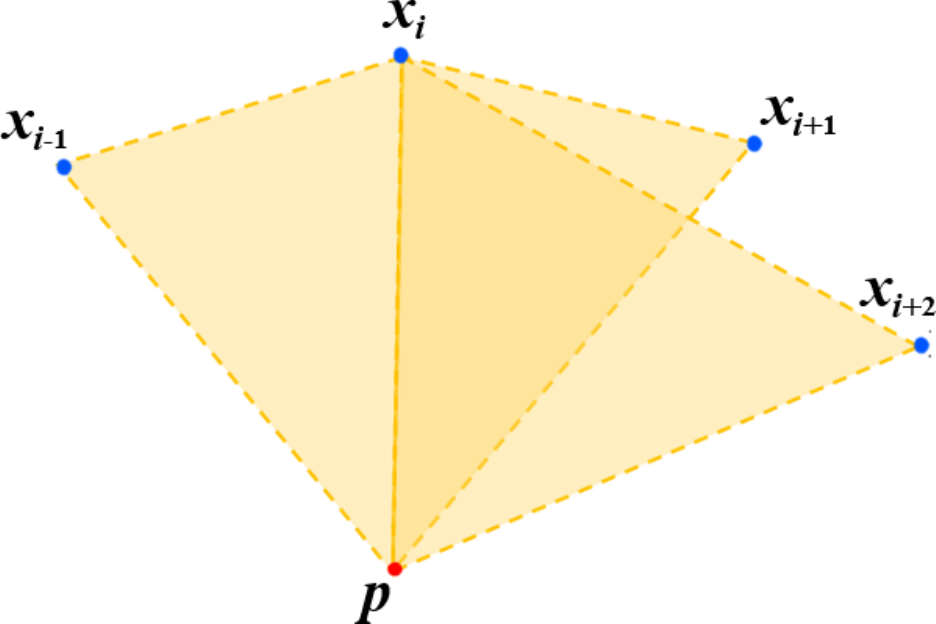}
\caption{Distorted surface caused by 3 triangles.}
\label{fig:nonmanifold}
\end{wrapfigure}
\textbf{Number of Local Surfaces}.
Local triangle surfaces used in~\cref{sec:risp} are the base for rotation invariant surface property (RISP) construction. Thus, it is important to analyse the effects of different number of surfaces. 
We conduct performance tests by setting the number of local surfaces to $1$, $2$, $3$, and $4$, resulting in overall accuracies of $94.7$, $96.0$, $91.9$, and $89.8$ respectively.
It worth noting that for more surfaces, we can define more RISPs. From the results, we see the accuracy drops a bit when single surface is constructed. This is because smaller area cannot capture sufficient surface structures and less effective features can lower the performance. When we increase the number of surfaces to 3 and 4, the performance decreases. This shows that more surfaces can result in inferior performance because we construct triangle surfaces just to approximate the true surface. More surfaces may mess up the surface structure and affect the performance. Take~\cref{fig:nonmanifold} for example, suppose we are constructing the local surfaces ($\triangle px_{i}x_{i+1}$ $\triangle px_{i}x_{i-1}$) with regard to the neighbor point $x_{i}$, adding one more triangle $\triangle px_{i}x_{i+2}$ will result in distorted/non-manifold surfaces. Thus, we set number of surfaces as 2 throughout the paper. 

\section{Discussion}
\textbf{Limitation.}
Though effective, \ourconv may require longer training time (e.g. more epochs) to converge due to large number of training parameters (\cref{tab:classification_modelnet40}). This is a common issue of transformer. But this does not affect the inference speed as shown in the ablation study (\cref{tab:parameters}).

\textbf{Conclusion.}
We have presented a new framework for achieving rotation invariance on 3D point cloud. We construct local triangle surfaces to better represent the local surface structures based on which we design highly expressive rotation invariant surface properties. We integrate the properties into an architecture named \ourconv to extract refined rotation invariant features. Based on RISurConv, we finally build up effective rotation invariant neural networks for 3D point cloud classification and segmentation with supreme performance achieved not only closing the performance gap with non-rotation invariant approaches but also surpassing the state-of-the-art methods by a large margin. 

\section*{Acknowledgements}
This research is supported by the Singapore Ministry of Education (MOE) Academic Research Fund (AcRF) Tier 1 grant (MSS23C010), Ningbo 2025 Science \& Technology Innovation Major Project (No. 2022Z072, No.2023Z044), and Key Research \& Development Plan of Zhejiang Province (2024C01017).

%
%
\bibliographystyle{splncs04}
\bibliography{main}

\newpage

\appendix
\section*{Supplementary Material}
\section{Overview}
In this document, we provide the mathematical proof for the completeness of our Rotation Invariant Surface Properties (RISP) defined in the main paper. 
Additionally, We show the output dimensions of each layer of the proposed neural network and explain the deconvolution layer in more details. 
The performance of the proposed method on other baselines is also investigated.
We also evaluate our method on another large-scale semantic segmentation S3DIS dataset. Finally, we look into the classification performance by comparing the per-class accuracies with the state-of-the-art methods.

\section{Completeness Proof of RISP}
\label{proof}
\begin{figure}[htb]
\centering
\includegraphics[width=0.6\linewidth]{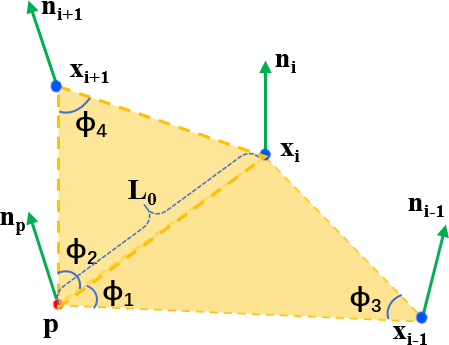}
\caption{RISP for two triangles. $p$ is the reference point, while $x_{i}$, $x_{i-1}$ and $x_{i+1}$ are the two neighbors. $n_{p}$, $n_{i}$, $n_{i-1}$ and $n_{i+1}$ are the corresponding normal vectors located at $p$, $x_{i}$, $x_{i-1}$, and $x_{i+1}$ respectively. }
\label{fig:two-triangles}
\end{figure}

In this section, we provide the completeness proof of RISP comprising 14 properties defined in the main paper. To facilitate illustration, we write down the RISP Equation here again:
\begin{equation}
\begin{split}
\label{suppl:eq:1}
	\mathrm{RISP}(x_i) = [L_{0}, \phi_{1}, \phi_{2}, \phi_{3}, \phi_{4}, \phi_{5}, \\ \alpha_{1}, \alpha_{2},  \beta_{1}, \beta_{2}, \theta_{1}, \theta_{2}, \gamma_{1}, \gamma_{2}]\,
\end{split}
\end{equation}
where $L_{0}$ is the distance from reference $p$ to neighbor $x_{i}$, while  $\phi_{1}$ to $\phi_{5}$ measure the two triangles as well as the relationship between the two triangle surfaces with regard to the edge $\vv{px_{i}}$ in the Euclidean space:
\begin{align} 
\label{suppl:eq:2}
\phi_{1} &= \angle\left( \vv{x_{i-1}p} , \vv{x_{i}p} \right), \\
\phi_{2} &= \angle\left( \vv{x_{i+1}p}, \vv{x_{i}p} \right), \nonumber\\
\phi_{3} &= \angle\left( \vv{x_{i-1}x_{i}}, \vv{x_{i-1}p} \right), \nonumber\\
\phi_{4} &= \angle\left(  \vv{x_{i+1}p}, \vv{x_{i+1}x_{i}} \right), \nonumber \\
\phi_{5} &= \angle\left(  \vv{x_{i+1}p} \times \vv{x_{i}p}, \vv{x_{i-1}p} \times \vv{x_{i}p} \right). \nonumber 
\end{align}
while other properties describe the two surfaces in the tangent space, e.g. normal vectors can define the directions in which the surface is bending away from the tangent space:

\begin{align} 
\label{suppl:eq:3}
\alpha_{1} &= \angle\left( \vv{n_{p}} , \vv{x_{i}p} \right),  \alpha_{2} = \angle\left( \vv{n_{p}}, \vv{x_{i-1}p} \right),  \\ 
\beta_{1} &= \angle\left( \vv{n_{i}} , \vv{x_{i}p} \right), \beta_{2} = \angle\left( \vv{n_{i}}, \vv{x_{i-1}x_{i}} \right), \nonumber \\
\theta_{1} &= \angle\left( \vv{n_{i-1}} , \vv{x_{i-1}p} \right), \theta_{2} = \angle\left( \vv{n_{i-1}}, \vv{x_{i-1}x_{i}} \right), \nonumber \\
\gamma_{1} &= \angle\left( \vv{n_{i+1}}, \vv{x_{i+1}x_{i}} \right), \gamma_{2} = \angle\left( \vv{n_{i+1}}, \vv{x_{i+1}p}  \right). \nonumber
\end{align}

To prove the completeness of these 14 properties, we start from single triangle analysis, and then extend to two triangles. 

\subsection{Completeness of RISP for One Triangle}
\begin{figure}[t]
\centering
\includegraphics[width=0.65\linewidth]{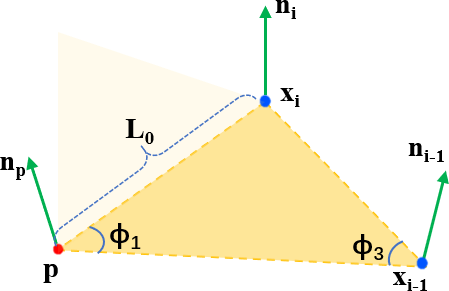}
\caption{RISP analysis for single triangle. }
\label{fig:one-triangle}
\end{figure}

Take Figure~\ref{fig:one-triangle} for instance, $L_{0}$ of Equation~\ref{suppl:eq:1} with $\phi_{1}$ and $\phi_{3}$ of Equation\ref{suppl:eq:2} can uniquely solve a triangle. We prove this statement as follows.

Given the fact that the sum of the angles in a triangle is always 180 degrees, the third angle can always be determined.

Based on the Law of Sines formula, we can find the lengths of the other two sides of the triangle. The Law of Sines states that in any triangle ABC with sides a, b, and c and opposite angles A, B, and C, the following relationship holds:
\begin{equation}
\label{eq:4}
\frac{a}{sin(A)} = \frac{b}{sin(B)} = \frac{c}{sin(C)}. 
\end{equation}
By using these formulas, we can calculate all the missing angles and side lengths of the triangle, and thus uniquely solve it. To prove the completeness of $L_{0}$, $\phi_{1}$ and $\phi_{3}$ for describing a triangle in Euclidean space, we need to show that any other triangle with the same angles and edge length must be congruent (i.e., they have the same shape and size).

Now, consider two triangles with the same two angles $\phi_{1}$ and $\phi_{3}$ and the same edge length $L_{0}$. Let $b_1$ and $c_1$ be the lengths of the other two edges of the first triangle, and let $b_2$ and $c_2$ be the lengths of the other two edges of the second triangle. We have already shown that $b_1 = b_2$ and $c_1 = c_2$, so the two triangles have the same shape.

To show the two triangles also have the same size, we can use the formula for the area of a triangle in terms of its edges:
\begin{equation}
\label{eq:5}
A = \frac{1}{2}L_{0}b \sin \phi_{1} = \frac{1}{2}bc \sin \phi_{3}
\end{equation}

Since $L_{0}$, $\phi_{1}$ and $\phi_{3}$ are the same for both triangles, we have $A_1 = A_2$ if and only if $b_1 = b_2$ and $c_1 = c_2$. Therefore, the two triangles must be congruent. This shows that two angles and one edge length are complete for describing a triangle in Euclidean space, up to congruence.

While $L_{0}$, $\phi_{1}$ and $\phi_{3}$ determine the triangle in the Euclidean space uniquely, the properties in tangent space is also crucial for analysing the behaviour of the surface.
The normal vector defines the direction in which the surface is bending away from the tangent space at that point. So, we use $\alpha_{1}$, $\alpha_{2}$, $\beta_{1}$, $\beta_{2}$, $\theta_{1}$, $\theta_{2}$ of Equation~\ref{suppl:eq:3} as the rotation invariant properties to encode the angles between the normals and the triangle edges. To prove the completeness of these 6 properties for describing the relationship between the normals and the triangle, we use $\theta_1$ and $\theta_2$ for illustration which encode the angles between $n_{i-1}$ and two edges located at $x_{i-1}$. To prove the completeness, we need to show that any other normal vector at the same point ($x_{i-1}$) on the triangle can be obtained by rotating the original normal vector by some combination of the two edges of the triangle, with angles equal to $\theta_1$ and $\theta_2$.

Let the original normal vector be denoted by $\vv{n_{i-1}}$ for this case, and let $\vv{e_1}$ and $\vv{e_2}$ be the two edges of the triangle at the same point as $\vv{n_{i-1}}$.

Consider any other normal vector $\vv{n_{i-1}}'$ at the same point on the triangle. We can express $\vv{n_{i-1}}'$ as a linear combination of $\vv{e_1}$, $\vv{e_2}$, and $\vv{n_{i-1}}$, as follows:
\begin{equation}
\label{eq:6}
\vv{n_{i-1}}' = a\vv{e_1} + b\vv{e_2} + c\vv{n_{i-1}}
\end{equation}

where $a$, $b$, and $c$ are scalar coefficients.

We can then express $a$, $b$, and $c$ in terms of the angles between $\vv{n_{i-1}}'$ and the two edges of the triangle, $\theta_1'$ and $\theta_2'$, as follows:

$$a = \frac{\vv{n_{i-1}}' \cdot \vv{e_1}}{|\vv{e_1}| |\vv{n_{i-1}}'|} = \frac{\cos \theta_1'}{\sqrt{1-\cos^2 \theta_1'}} |\vv{n_{i-1}}'|$$

$$b = \frac{\vv{n_{i-1}}' \cdot \vv{e_2}}{|\vv{e_2}| |\vv{n_{i-1}}'|} = \frac{\cos \theta_2'}{\sqrt{1-\cos^2 \theta_2'}} |\vv{n_{i-1}}'|$$

$$c = \sqrt{1-a^2-b^2}$$

We can then express $\vv{n_{i-1}}'$ in terms of the original normal vector and the two edges of the triangle as follows:
\begin{equation}
\begin{split}
\label{eq:7}
\vv{n_{i-1}}' = \cos \theta_1' \frac{\vv{n_{i-1}} \times \vv{e_1}}{|\vv{n_{i-1}} \times \vv{e_1}|} + \cos \theta_2' \frac{\vv{n_{i-1}} \times \vv{e_2}}{|\vv{n_{i-1}} \times \vv{e_2}|} + \\ \sqrt{1-\cos^2 \theta_1' - \cos^2 \theta_2'} \vv{n_{i-1}}
\end{split}
\end{equation}

This expression shows that any other normal vector at the same point on the triangle can be obtained by rotating the original normal vector by some combination of the two edges of the triangle, with angles equal to $\theta_1$ and $\theta_2$. Therefore, the two angles between a normal vector at one point of the triangle and the two edges of that triangle are complete in describing the relationship between the normal vector and the triangle.
\subsection{Completeness of RISP for Two Triangles}
Now we prove the completeness of RISP for two triangles. As is shown in Figure~\ref{fig:two-triangles}, by adding one more point $x_{i+1}$, we have two triangles sharing the edge $\vv{x_{i}p}$. We need more properties to encode the second triangle as well we the relationship between these two triangles. 

We use $L_{0}$, $\phi_{2}$ and $\phi_{4}$ to uniquely determine the second triangle in the Euclidean space. We also we use $\phi_{5}$ to encode the information about the relative orientation of the two triangles. In the tangent space, we add two properties $\gamma_{1}$ and $\gamma_{2}$ to encode the angles between $n_{i+1}$ and the two adjacent edges. 

Note that we do not explicitly encode the angle between $n_{i}$ and $\vv{x_{i}x_{i+1}}$ and the angle between $n_{p}$ and $px_{i+1}$, because these two angles can be computed based on the RISP properties.

The relative position of $n_{i}$ to the first triangle $\triangle px_{i}x_{i-1}$ is already fixed by $\beta_{1}$ and $\beta_{2}$, while the relative position of the second triangle $\triangle px_{i}x_{i+1}$ to the first triangle $\triangle px_{i}x_{i-1}$ is also fixed by $\phi_{5}$. So, the angle between $n_{i}$ and $x_{i}x_{i+1}$ is also fixed. 

To compute the angle $\mu$ between $n_{i}$ and $\vv{x_{i}x_{i+1}}$ for example, we can use the following law of tetrahedron to find the angle:

$$\cos \mu = \cos\phi_{6}\cos \beta_{1} + \sin\phi_{6}\sin \beta_{1} \cos \phi_{5},$$
where $\phi_{6}=\pi - \phi_{2}- \phi_{4}$

Now, we have explained all the 14 properties used in Equation~\ref{suppl:eq:1} and proved the completeness. In the ablation study of the main paper, we also see that when the angle $\mu$ between $n_{i}$ and $\vv{x_{i}x_{i+1}}$ is added the performance does not improve. This also verifies the completeness of RISP. 

\section{\ourconv Network}
In Figure 3 of the main paper, we have plotted the \ourconv network architecture. To further illustrate the network details, we list the output dimensions and points for each layer of the classification and segmentation as follows in Table~\ref{tab:network}. Please note that the deconvolution process of the segmentation network follows a similar approach to \ourconv. The key distinction lies in our convolution's output, which targets a point subset with an increased number of feature channels. In contrast, deconvolution directs its output to a point set with more points compared to the input, yet with fewer feature channels. Specifically, if we consider deconvolution starting with the set of $N_l$ points at layer $l$, the \ourconv operator is applied to upsample the features of these $N_l$ points to a set of $N_{l+1}$ points at the subsequent layer $l+1$. This deconvolution process iterates until the point cloud regains its original number of points, denoted as $N$.

It's important to highlight that, due to the availability of point subsets during the downsampling phase in the encoder section, there is no need to generate new points during the upsampling. Instead, we can efficiently reuse these subsets and propagate the features through interpolation.
\begin{table}[tb]
   \caption{The details of our network. Here, $K$ represents the number of categories, and $N$ represents the number of input points.}
    \label{tab:network}
    \centering
    \begin{tabular}{l | c} 
        \noalign{\smallskip}\hline\noalign{\smallskip}
        Module & Output shape ($dims \times points$)\\
        \noalign{\smallskip}\hline\noalign{\smallskip}
        \textbf{Classification / Retrieval} & \\
        Input tensor & $3 \times 1024$ \\
        \ourconv & $32 \times 1024$ \\
        \ourconv & $64 \times 512$ \\
        \ourconv & $128 \times 256$ \\
        \ourconv & $256 \times 128$\\
        \ourconv & $512 \times 1$\\
        Transformer Encoder & $512 \times 1$ \\
        Fully connected & $256 \times 1$ \\
        Fully connected & $128 \times 1$ \\ 
        Softmax & $K \times 1$ \\
        \\
        \textbf{Segmentation} & \\
        Input tensor & $3 \times N$ \\
        \ourconv & $64 \times 512$ \\
        \ourconv & $128 \times 256$ \\
        \ourconv & $256 \times 128$ \\
        \ourconv & $512 \times 64$
        \\
        \ourconv & $512 \times 128$ \\
        Skip connection & $768 \times 128$ \\
        MLP & $512 \times 128$ \\
        \ourconv & $512 \times 256$ \\
        Skip connection & $640 \times 256$ \\
        MLP & $256 \times 256$ \\
        \ourconv & $256 \times 512$ \\
        Skip connection & $320 \times 512$ \\
        MLP & $128 \times 512$ \\
        \ourconv & $K \times N$ \\
        \noalign{\smallskip}\hline\noalign{\smallskip}
    \end{tabular}
\end{table}

\section{Other Baselines}
\label{baseline}
Our network architecture is flexible in that it can be used for different baselines by replacing the xyz input with the proposed rotation invariant properties. Use pointnet++ and DGCNN as baselines, we tested the performance and the accuracies improve from 28.6\% to 90.2\% and 20.6\% to 92.0\% respectively on the ModelNet40 dataset under the challenging z/SO3 mode.

\section{Large-Scale Scene Segmentation on S3DIS}
\label{semantic}
To evaluate the performance on semantic segmentation, we further conduct an experiment on S3DIS dataset (Armeni et al. 2016) which is a large-scale indoor scene dataset comprising 3D scans from Matterport scanners in 6 indoor areas including 271 rooms with each point is annotated with one of the semantic labels from 13 categories. In this experiment, we use Area-5 for testing to better measure the generalization ability, and report the results in SO3/SO3 and z/SO3 scenario. 
Different from previous works that rotate the whole scan, we only rotate the instances such that it is closer to the real-life situation. We compare our method with traditional 3D deep learning methods such PointNet~\cite{qi2017pointnet}, PointCNN~\cite{li2018pointcnn} and PointTransforer~\cite{zhao2021point} as well as rotation invariant methods such as RIConv~\cite{zhang-riconv-3dv19}, GCAConv~\cite{zhang2020global}, and RIConv++~\cite{zhang2022riconv++}. The results are shown in Table~\ref{tab:s3dis}. From the results, we can see that our method outperforms RIConv++ by 2.1\% mIoU, and works consistently for both rotation scenarios. Note that we use the original source codes of these methods and do not modify them by add normal computation. Thus, only RIConv++ and our method provide two versions for input with normal and without normal in this experiment.

\begin{table}[t]
	\caption{Comparisons of the semantic segmentation accuracy (mIoU, \%) on the S3DIS dataset (Area-5).}
\label{tab:s3dis}
	\centering
	\begin{tabular}{l|cccc}
		\noalign{\smallskip}\hline\noalign{\smallskip}
		Method   & SO3/SO3 & z/SO3\\
		\noalign{\smallskip}\hline\noalign{\smallskip}
		PointNet         & 39.1 & 31.2 \\
		PointCNN         & 51.2 & 30.6 \\
		PointTransformer & 53.6 & 44.8 \\
		\noalign{\smallskip}\hline\noalign{\smallskip}
		RIConv                &53.8 &53.8 \\
		GCAConv               &54.3 &54.1 \\
		RIConv++ (w/o normal) & 57.8 & 57.8\\
		RIConv++ (w/ normal)  & 58.0 & 58.0 \\
		\noalign{\smallskip}\hline\noalign{\smallskip}
		Ours (w/o normal)     & \underline{59.3} & \underline{59.3}   \\
		Ours (w/ normal)      & \textbf{60.1} & \textbf{60.1} \\
		\noalign{\smallskip}\hline
	\end{tabular}
\end{table}

\section{Per-Class Accuracies}
\label{per-class}

We further evaluate the accuracy of the classification task per object class under the z/SO3 mode. 
The per-class accuracies are shown in Table~\ref{tab:modelnet40_perclass_v2}. Comparing to the results by non rotation invariant methods like PointNet ~\cite{qi2017pointnet}, PointNet++~\cite{qi2017pointnet++} and PointCNN ~\cite{li2018pointcnn}, as well as rotation invariant methods like RIConv~\cite{zhang-riconv-3dv19}, GCAConv~\cite{zhang2020global}, and RIConv++~\cite{zhang2022riconv++}, we can see that our method outperforms other methods on most classes with a large margin. Our method achieves 1st (in bold) in 36 out of 40 classes, while RIConv~\cite{zhang-riconv-3dv19}, GCAConv~\cite{zhang2020global}, and RIConv++~\cite{zhang2022riconv++} only have  2, 9, 9 classes achieving 1st, respectively.
\begin{table}[htp]
\caption{Per-class accuracy of object classification in z/SO3 scenario with the ModelNet40 dataset.}
	\label{tab:modelnet40_perclass_v2}
	\begin{center}
	\begin{tabular}{l| p{28pt}p{28pt}ccccp{30pt}p{28pt}}
		\noalign{\smallskip}\hline\noalign{\smallskip}
		Network  & aero & bathtub & bed  & bench & bookshelf & bottle & bowl  & car  \\ 
		\noalign{\smallskip}\hline\noalign{\smallskip}
		PointNet \cite{qi2017pointnet}  &12.0 &2.0 & 8.0 & 10.0 & 15.0 &14.0&5.0 &12.0\\ 
		
		PointNet++ \cite{qi2017pointnet++} & 53.0  &  2.0  &  18.0  &  10.0  &  29.0  &  22.0  & 20.0  &  13.0  \\ 
		PointCNN \cite{li2018pointcnn} & 60.0 & 10.0 & 20.0 & 10.0 & 20.0 & 37.0 & 25.0 & 34.0   \\ 
		
		RIConv \cite{zhang-riconv-3dv19} &\textbf{100.0} &82.0 &94.0 &\textbf{80.0} &93.0 &94.0 &100.0 &98.0 \\
		GCAConv \cite{zhang2020global} & \textbf{100.0} & 90.0 & 98.0 &\textbf{80.0} & 95.0 &97.0 & \textbf{100.0} & 98.0 \\
		RIConv++ \cite{zhang2022riconv++} &\textbf{100.0}   &90.0  &97.0  &\textbf{80.0}  &\textbf{97.0}  &94.0  &\textbf{100.0}  &\textbf{100.0} \\
		Ours &\textbf{100.0} & \textbf{98.0} &\textbf{99.0} &\textbf{80.0} & \textbf{97.0} & \textbf{99.0} &95.0 &\textbf{100.0} \\
		\noalign{\smallskip}\hline\noalign{\smallskip}
		& chair & cone & cup & curtain & desk & door & dresser & flower pot\\
		\noalign{\smallskip}\hline\noalign{\smallskip}
		PointNet\cite{qi2017pointnet}  &9.0&15.0 &0.0&0.0&16.3&5.0&8.1&0.0\\
		PointNet++ \cite{qi2017pointnet++}  &  32.0  &  20.0  &  15.0  &  45.0  & 2.3  & 30.0  &  9.3  & 15.0    \\
		PointCNN \cite{li2018pointcnn}  & 46.0  & 25.0  & 15.0  & 40.0 & 34.9 & 30.0 & 32.6 & 25.0 \\  
		
		RIConv \cite{zhang-riconv-3dv19} &96.0 &90.0 & 60.0 & 95.0 & 79.1 & 85.0 & 73.3 & 30.0  \\
		GCAConv \cite{zhang2020global} & \textbf{98.0} & 90.0 & 55.0 &95.0 & 81.4 &80.0 & 68.6 & 10.0 \\
		RIConv++ \cite{zhang2022riconv++} &97.0   &95.0    &70.0   &\textbf{100.0}   & 83.7   & 85.0   &72.1  &15.0 \\
		Ours & \textbf{98.0} & \textbf{100.0} & \textbf{80.0} & \textbf{100.0} & \textbf{95.3} & \textbf{95.0} &\textbf{93.0} & \textbf{70.0} \\
		\noalign{\smallskip}\hline\noalign{\smallskip}
		& glass box & guitar & keyboard & lamp & laptop & mantel & monitor & night stand\\
		\noalign{\smallskip}\hline\noalign{\smallskip}
		PointNet \cite{qi2017pointnet}   &4.0&36.0&5.0&15.0 &15.0&4.0&11.0&3.5\\
		PointNet++ \cite{qi2017pointnet++}  & 11.0  & 47.0  & 50.0  & 10.0    & 15.0 &  10.0  & 36.0  &  1.2  \\
		PointCNN \cite{li2018pointcnn}  & 35.0 & 46.0 & 50.0 &  20.0  & 20.0 & 38.0 & 35.0 & 40.7\\ 
		RIConv \cite{zhang-riconv-3dv19} &96.0 &99.0 &95.0 & 80.0 &95.0 & 91.9 & 97.0 & 77.9 \\
		GCAConv \cite{zhang2020global} & \textbf{97.0} & \textbf{100.0} & 95.0 &\textbf{85.0} & \textbf{100.0} &93.0 & 98.0 & 73.3 \\
		RIConv++ \cite{zhang2022riconv++} &\textbf{97.0}   &100.0   &\textbf{100.0}   &95.0   &100.0  & 94.0   &96.0   &80.2 \\
		Ours & 96.0 & \textbf{100.0} & \textbf{100.0} & \textbf{85.0} &\textbf{100.0} & \textbf{99.0} & \textbf{99.0} & \textbf{83.7} \\
		\noalign{\smallskip}\hline\noalign{\smallskip}
		 & person & piano & plant & radio & range hood & sink & sofa & stairs \\
		\noalign{\smallskip}\hline\noalign{\smallskip}
		PointNet \cite{qi2017pointnet}    &5.0&36.7&55.0&5.0&4.0&20.0&11.0&25.0 \\
		PointNet++ \cite{qi2017pointnet++}   & 20.0  &  5.0  & 71.0 &  20.0   & 9.0  &  5.0 & 21.0  & 10.0 \\
		PointCNN \cite{li2018pointcnn}  & 15.0 & 34.0 & 26.0  & 10.0 &  28.0  & 20.0   & 32.0 &  30.0\\ 
		RIConv \cite{zhang-riconv-3dv19} &85.0 &90.8 & 83.0 &55.0 & 87.0 &75.0 & 92.0 &85.0 \\
		GCAConv \cite{zhang2020global} & 90.0 & 91.0 & \textbf{93.0} &65.0 & 86.0 &70.0 & 93.0 & 80.0 \\
		RIConv++ \cite{zhang2022riconv++} &\textbf{100.0}   &94.0   & 91.0   & 65.0   &95.0    &75.0   &97.0   &95.0 \\
		Ours   & \textbf{100.0} & \textbf{97.0} & 82.0 & \textbf{95.0} & \textbf{96.0} & \textbf{85.0} &\textbf{98.0} &\textbf{95.0} \\
		\noalign{\smallskip}\hline\noalign{\smallskip}
		& stool & table & tent & toilet & tv stand & vase & wardrobe & xbox\\
		\noalign{\smallskip}\hline\noalign{\smallskip}
		PointNet \cite{qi2017pointnet} &5.0&3.0&5.0&20.0&4.0&26.3&0.0&10.0 \\
		PointNet++ \cite{qi2017pointnet++}  & 10.0  &  9.0  & 15.0  & 13.0 & 2.0  & 85.0 &15.0 &  20.0 \\
		PointCNN \cite{li2018pointcnn}	 &  20.0 &  36.0 &  15.0 &  33.0  & 29.0 & 70.0   &40.0 &  15.0\\
		RIConv \cite{zhang-riconv-3dv19} &60.0 &80.0 &70.0 &95.0 &78.0 &76.8 &70.0 &65.0\\
		GCAConv \cite{zhang2020global} & 75.0 & 84.0 & 95.0 &99.0 & 81.0 &77.0 & \textbf{70.0} & 75.0 \\
		RIConv++ \cite{zhang2022riconv++} &85.0    &81.0    &95.0    &99.0    &87.0    &84.0    &45.0    &90.0 \\
		Ours  & \textbf{90.0} & \textbf{95.0} & \textbf{100.0} & \textbf{100.0} & \textbf{95.0} &\textbf{93.0} & 55.0 & \textbf{85.0}\\
		\noalign{\smallskip}\hline\noalign{\smallskip}
	\end{tabular}
	\end{center}
\end{table}
\end{document}